\documentclass{article}

\usepackage{graphicx}
\usepackage{color}
\usepackage{subfig}
\usepackage{caption}

\usepackage{microtype}
\usepackage{graphicx}
\usepackage[utf8]{inputenc} 
\usepackage[T1]{fontenc}    
\usepackage{url}            
\usepackage{booktabs}       
\usepackage{amsfonts}       
\usepackage{nicefrac}       
\usepackage{microtype}      
\usepackage{natbib}
\usepackage{amsmath}        
\usepackage{gensymb}

\usepackage{hyperref}

\usepackage[accepted]{icml2019}
    
\icmltitlerunning{LiDAR Sensor modeling and Data augmentation with GANs for Autonomous driving}

\begin{document}

\twocolumn[
\icmltitle{LiDAR Sensor modeling and Data augmentation \\ with GANs for Autonomous driving}
           


\icmlsetsymbol{equal}{*}

\begin{icmlauthorlist}
\icmlauthor{Ahmad El Sallab}{equal,va}
\icmlauthor{Ibrahim Sobh}{equal,va}
\icmlauthor{Mohamed Zahran}{equal,va}
\icmlauthor{Nader Essam}{equal,va}
\end{icmlauthorlist}

\icmlaffiliation{va}{Research Team, Valeo Egypt}

\icmlcorrespondingauthor{Mohamed Zahran}{mohamed.zahran@valeo.com}

\icmlkeywords{Sensor Modeling, Data augmentation, Generative adversarial networks, LiDAR}

\vskip 0.3in
]



\printAffiliationsAndNotice{\icmlEqualContribution} 


%




\begin{abstract}
In the autonomous driving domain, data collection and annotation from real vehicles are expensive and sometimes unsafe. Simulators are often used for data augmentation, which requires realistic sensor models that are hard to formulate and model in closed forms. Instead, sensors models can be learned from real data. The main challenge is the absence of paired data set, which makes traditional supervised learning techniques not suitable. In this work, we formulate the problem as image translation from unpaired data and employ CycleGANs to solve the sensor modeling problem for LiDAR, to produce realistic LiDAR from simulated LiDAR (sim2real).
Further, we generate high-resolution, realistic LiDAR from lower resolution one (real2real). The LiDAR $3$D point cloud is processed in Bird-eye View and Polar $2$D representations. The experimental results show a high potential of the proposed approach.
\end{abstract}

\section{Introduction}
\label{sec: introduction}
Environment perception is crucial for safe Automated Driving. While a variety of sensors powers low-level perception, high-level scene understanding is enabled by computer vision, machine learning, and deep learning, where the main challenge is in data scarcity. On the one hand, deep learning models are known for their hunger to data, on the other hand, it is both complex and expensive to obtain a massively labeled dataset, especially for non-common sensors, like LiDAR, Radar and UltraSonic sensors.

Data augmentation techniques could be used to increase the amount of the labeled data artificially. As simulators and game engines become more realistic, they become an efficient source to obtain annotated data for different environments and different conditions. However, there remain two main challenges: 1) For non-common sensors, we need to develop a model that perceives the simulator environment and produce data in the same way the physical sensor would do, which we call "Sensor Model" and 2) The sensor model should consider the realistic noisy conditions in the natural scenes, unlike the perfect simulation environments, such as imperfect reflections, and material reflectivity for LiDAR sensors as an example. Sensors models are hard to develop in a closed mathematical form. Instead, they better be learned from real data. However, it is almost impossible to get a paired dataset, composed of noisy and clean data for the same scene, because this requires to have the sensor model in the first place. This dilemma calls for the need to unsupervised methods, which can learn the model from unpaired and unlabeled data.

In this work, we formulate the problem of sensor modeling as an image-to-image translation and employ CycleGAN~\cite{Zhu-ICCV-2017} to learn the mapping. 
We formulate the two domains to translate between 1) the real LiDAR point cloud (PCL) and 2) the simulated LiDAR point cloud. In another setup, we also translate low-resolution LiDAR, into higher resolution generated LiDAR. For the realistic LiDAR, we use the KITTI dataset \cite{geiger2013vision}, and for the simulated LiDAR, we use the CARLA simulator \cite{dosovitskiy2017carla}. The processing of the input LiDAR is done in two ways 
\begin{itemize}
\item Bird-eye View ($2$D BEV): projection as an Occupancy-Grid Map (OGM)
\item Polar-Grid Map ($2$D PGM): where the LiDAR PCL is transformed in the polar coordinates, which maps precisely the physical way the LiDAR sensor scans the environment. Accordingly, no losses occur where all points in the PCL are considered. Furthermore, the reverse mapping from the $2$D PGM to the $3$D PCL is possible.
\end{itemize}

The rest of the paper is organized as follows: first, the related work is presented, then the approach is formulated under the umbrella of CycleGANs, including the different translation domains. Following that, the experimental setup and the different environments and parameters are discussed. Finally, we conclude with the discussion and future works, focusing on the different ways to evaluate the generated data.
\section{Related Work}
\label{sec:related}
\label{RelatedWork}
Generative Adversarial Network (GAN) \cite{GAN} were proposed as a two-player minimax game between two neural networks, the generator and the discriminator. The generator tries to learn how to generate data samples as similar as possible to the real data samples. On the other hand, the discriminator is a binary classifier that tries to learn how to distinguish between the real samples from the generated (or fake) ones. This competitive game should end up at an equilibrium point at which the generator is able to produce realistic samples, while the discriminator is unable tell what is real and what is fake. The goal is to have the generated data distribution $p_g$ close to the real data distribution $p_d$. 
The generator $G$ is defined as a mapping function that maps the input noise $z$ to data space, and represented as a neural network with parameters $\theta_g$. The discriminator $D$, on the other hand, is a binary classifier represented as another neural network with parameters $\theta_d$.
Both $D$ and $G$ are trained simultaneously. The discriminator $D$ is trained to maximize the probability of assigning the correct label to either the real training samples or the fake generated samples, while the generator $G$ is trained to fool $D$ by minimizing the $\log(1 - D(G(z))$. Accordingly, both $D$ and $G$ are playing a two-player minimax game over a value function $V(G, D)$ defined as follows: 

\begin{eqnarray}
\label{GAN_loss}
    \min_{\theta_g}\max_{\theta_d} V(D, G)  = \mathbb{E}_{x \sim p_{\mathrm{data}(x)}} \big[ \log D_{\theta_d}(x) \big] \nonumber \\ 
     + \ \mathbb{E}_{z \sim p(z)} \Big[ \log\big(1 - D_{\theta_d}(G_{\theta_g}(z))\big) \Big] \;, \nonumber \\ 
\end{eqnarray}


In \textbf{image-to-image} translation the model is trained to map from images in a source domain to images in a target domain where conditional GANs (CGANs) are adopted to condition the generator and the discriminator on prior knowledge. Generally speaking, image-to-image translation can be either supervised or unsupervised. In the supervised setting, Pix2Pix~\cite{Isola-CVPR-2017}, SRGAN~\cite{ledig2017photo}, the training data is organized in pairs of input and the corresponding output samples. However, in some cases the paired training data is not available and only unpaired data is available. In this case the mapping is learned in an unsupervised way given unpaired data, as in CycleGAN~\cite{Zhu-ICCV-2017}, UNIT~\cite{liu2017unsupervised}.         
Moreover, instead of generating one image, it is possible to generate multiple images based on a single source image in multi-modal image translation. Multimodal translation can be paired Pix2PixHD~\cite{wang2018high}, BicycleGAN~\cite{zhu2017toward}, or unpaired MUNIT~\cite{huang2018multimodal}. Applying image-to-image translation for video results in the temporally incoherent output. A video synthesis approach based on GAN framework is proposed in~\cite{wang2018video}. As an example of $3$D data with GANs, Point Cloud GAN (PC-GAN)~\cite{li2018point} is proposed for learning to generate point clouds and to transform $2$D images into $3$D point clouds. A recent generator architecture for GAN based on style transfer approaches is proposed in \cite{karras2018style}, that enabled excellent control based on different levels of the generated images. GANs are used for domain adaptation to enable machine learning models trained on samples from a source domain to generalize to a target domain. For example, GraspGAN~\cite{bousmalis2017using} used pixel-level domain adaptation approach to reduce the number of real-world samples.

In this work, to enable LiDAR sensor modeling and data augmentation, unsupervised image-to-image translation is adapted to map between $2$D LiDAR domains where the complex $3$D point cloud processing is avoided. Two LiDAR domains are used; simulated LiDAR from CARLA urban car simulator and KIITI dataset.
\section{Approach}
\label{sec: Approach}
The main idea is to translate LiDAR from a source domain to another target domain. In the following experiments, two $2$D LiDAR representations are used. The first is the $2$D bird-view, where the $3$D point cloud is projected where the height view is lost. 
The second is the PGM method that represents the LiDAR $3$D point cloud into a $2$D grid that encodes both channels and the ray step angle of the LiDAR sensor. The $3$D point cloud can be reconstructed from the PGM $2$D representation given enough horizontal angular resolution. 

\subsection{Formulation and losses}
\begin{figure}[ht]
    \vskip 0.2in
    \begin{center}
        \fbox{\includegraphics[scale=0.185]{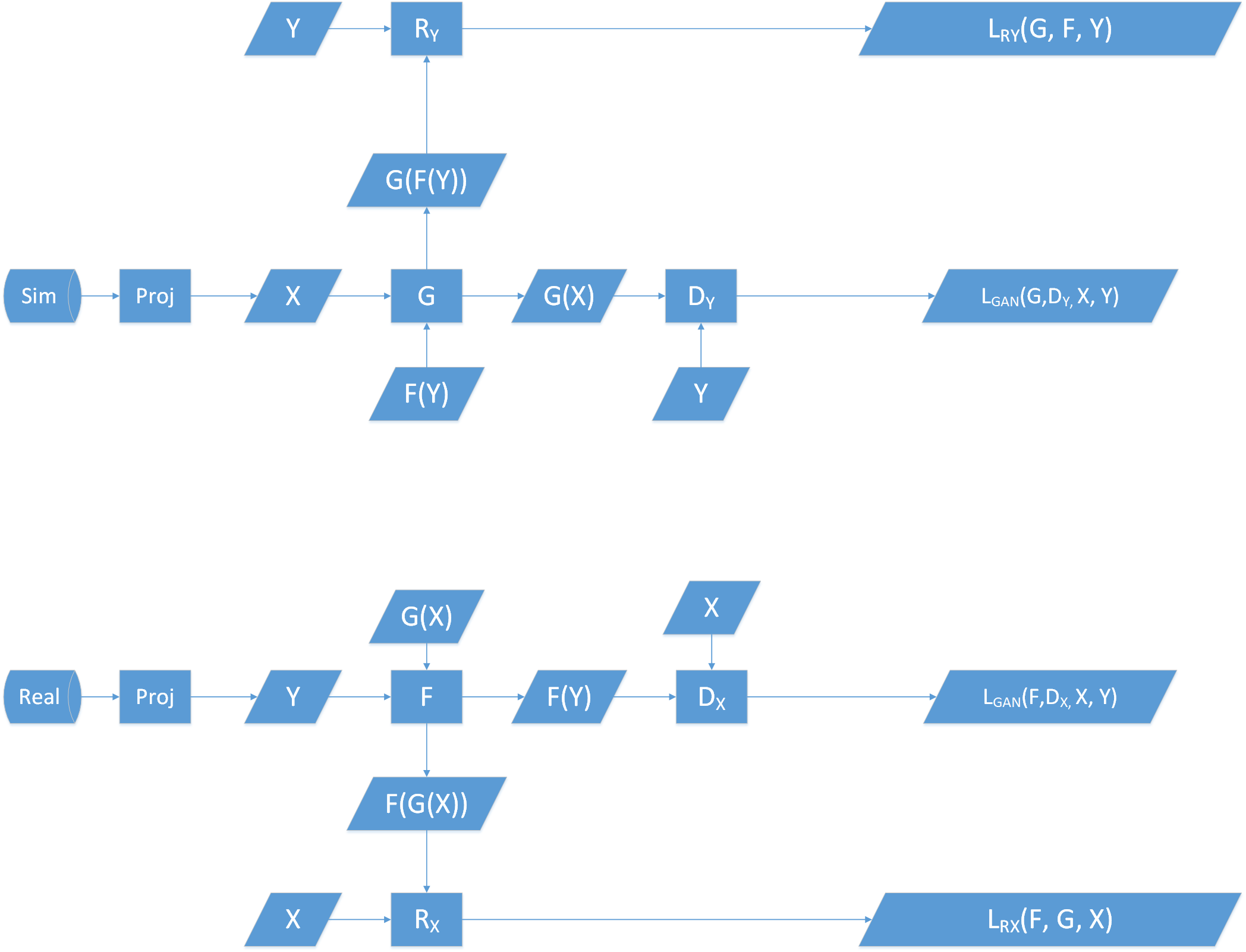}}
        \caption{CycleGAN model}
        \label{fig:vanilla_gan}
    \end{center}
\end{figure}

The full system overview is shown in figure \ref{fig:vanilla_gan}. Let $X$ be the simulated data, and $Y$ be the real data. Both signals correspond to the raw $3$D PCL projected on $2$D space. The projection techniques to be described later.
The $sim2real$ generator network is $G$, while the reverse $real2sim$ generator is $F$. The real data discriminator is $D_Y$, while the simulated data discriminator is $D_X$. The generative adversarial losses are as follows:

\begin{eqnarray}
\mathcal{L}_{GAN}(G, D_Y, X,Y) = \mathbb{E}_{y \sim{~} p_{data}(y)}\big[ \log D_Y(x) \big] \nonumber \\ 
 + \mathbb{E}_{x \sim{~} p_{data}(x)} \Big[ \log\big (1 - D_Y(G(x))\big) \Big] \;, \nonumber \\ 
\end{eqnarray}

\begin{eqnarray}
\mathcal{L}_{GAN}(F, D_X, X,Y) = \mathbb{E}_{x \sim{~} p_{data}(x)}\big[ \log D_X(x)\big]\nonumber  \\ 
 + \mathbb{E}_{y \sim{~} p_{data}(y)} \Big[ \log\big(1 - D_X(F(y))\big) \Big]\;, \nonumber \\ 
\end{eqnarray}

The main trick that enables CycleGANs to do the translation from unpaired data is the cycle consistency loss. Two cycle consistency losses are defined as follows:
\begin{eqnarray}
\mathcal{L}_{R_Y}(G, F, Y)\!\!\!\!\! &=& \!\!\!\!\!\mathbb{E}_{y \sim{~} p_{data}(y)}[R_Y] \nonumber\\ 
 \!\!\!\!\! & & \!\!\!\!\!=\  \mathbb{E}_{y \sim{~} p_{data}(y)}\Big[|| G(F(x)) - y||_1 \Big] \;, \nonumber \\ 
\end{eqnarray}
where: $R_Y = || G(F(x)) - y ||_1$

\begin{eqnarray}
\mathcal{L}_{R_X}(F, G, X) \!\!\!\!\! &=& \!\!\!\!\! \mathbb{E}_{x \sim{~} p_{data}(x)}[R_X] \nonumber\\ 
 \!\!\!\!\! & & \!\!\!\!\! = \ \mathbb{E}_{x \sim{~} p_{data}(x)}\Big[|| F(G(y)) - x ||_1 \Big] \;, \nonumber \\ 
\end{eqnarray}
where: $R_X = || F(G(y)) - x ||_1$

The overall loss function is as follows:
\begin{eqnarray}
\mathcal{L}(G,F,D_X,D_Y)  =
    \mathcal{L}_{GAN}(G, D_Y, X,Y) \nonumber\\ 
 + \ \mathcal{L}_{GAN}(F, D_X, X,Y) \nonumber\\ 
    + \ \lambda \Big[\mathcal{L}_{R_Y}(G, F, Y) + \mathcal{L}_{R_X}(F, G, X)\Big] \;, \nonumber \\ 
\end{eqnarray}

The $G$ and $F$ mapping functions are obtained by optimizing the overall loss as follows:

\begin{eqnarray}
G^*, F^* = \arg \min_{G, F} \max_{D_X, D_Y} \mathcal{L}(G,F,D_X,D_Y)
\end{eqnarray}

Note that, the following can be viewed as autoencoder reconstruction:
\begin{eqnarray}
G(F(Y)) = G \cdot F(Y) 
\end{eqnarray}
\begin{eqnarray}
F(G(X)) = F \cdot G(X) 
\end{eqnarray}

\subsection{LiDAR Translations}
\begin{figure}
        \centering
        \subfloat[]{\fbox{\includegraphics[scale=0.45]{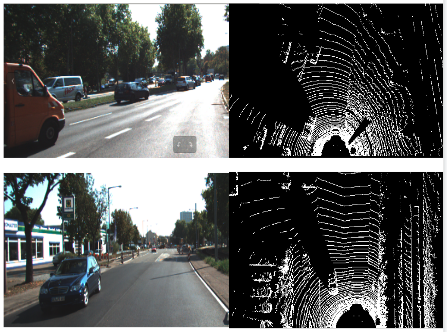}}}
        \centering  
        \vspace{1em}
        \subfloat[]{\fbox{\includegraphics[scale=0.45]{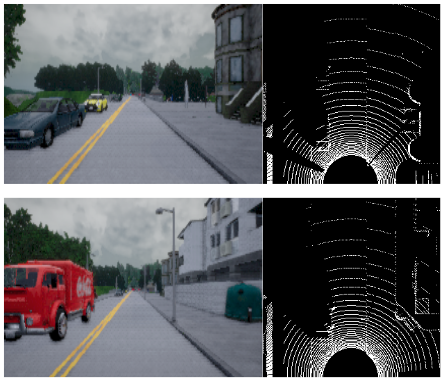}}}
    \caption{Sample RGB camera frame and the corresponding BEV frame. a: KITTI Dataset, b: Simulated CARLA Dataset.}
    \label{fig:datasets}
\end{figure}

Table ~\ref{tab:F1} shows the LiDAR specifications in each Dataset. Two datasets are considered; CARLA for the simulated domain ($X$) and KITTI for the real domain ($Y$). CARLA ~\cite{dosovitskiy2017carla} is an open source urban car simulator provides a highly realistic environment including rich urban towns with different layouts, surrounding trees, and buildings, different vehicles types and pedestrians. Moreover, CARLA has a configurable sensors suite including RGB cameras and basic LiDAR sensors, making it a high fidelity environment suitable for training and testing autonomous driving systems. The collected simulated data from the CARLA includes a large number of pedestrians and vehicles. 
KITTI ~\cite{geiger2013vision} is a publicly available raw dataset consists of sequenced frames. The dataset consists of 36 different annotated point cloud scenarios of variable length and a total of 12919 frames. These scenarios have diverse driving environments such as highway, roads, traffic, city and residential areas. They are also rich with dynamic and static objects. Samples from KITTI and CARLA dataset are shown in figure  \ref{fig:datasets}.
Translation from $2$D LiDAR is conducted between simulated CARLA and real KITTI data sets. This is an essential application where the simulated data is much easier to be collected compared to the real data. In this case, realistic data can be generated given simulated data which saves a lot of time and cost.  

\begin{table}[t]
  \caption{LiDAR specifications for all used datasets}
  \label{tab:F1}
  \vskip 0.15in
    \begin{center}
    \begin{small}
    \begin{sc}
    \begin{tabular}{lcccc}
    \toprule
    Name     & Layers     & H. FOV   & V. FOV & Range \\
    \midrule
    Carla   & 32/64  & 360.0\degree   & 44.0\degree  & 50m \\
    KITTI    & 64  & 360.0\degree   & 26.9\degree & 120m \\
    \bottomrule
\end{tabular}
\end{sc}
\end{small}
\end{center}
\vskip -0.1in
\end{table}

\section{Experiments}
\label{sec: experiments}
\begin{figure}
\centering
\fbox{\includegraphics[scale=0.27]{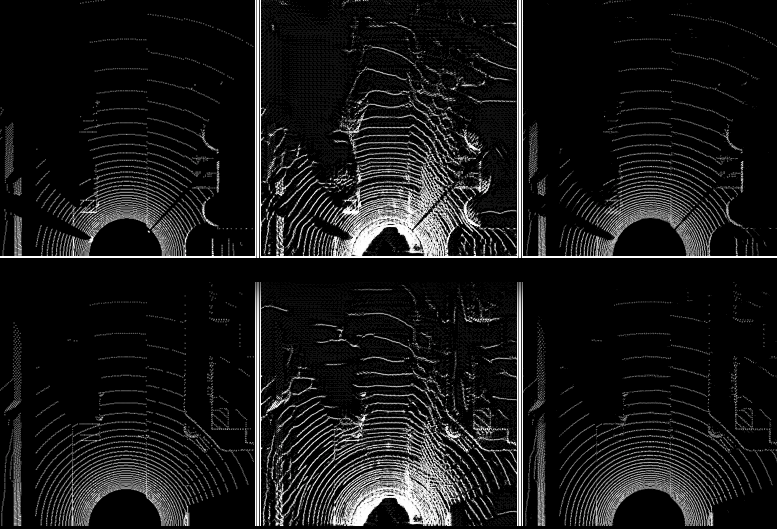}}
\caption{Left: Simulated Carla frame. Centre: Generated fake KITTI frame. Right: Reconstructed CARLA frame.}
\label{fig:C2K}
\end{figure}

\begin{figure}
\centering
\fbox{\includegraphics[scale=0.27]{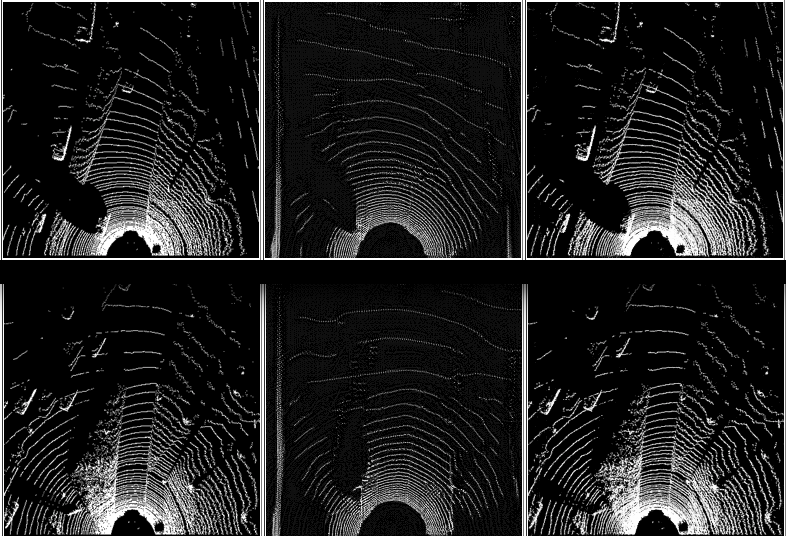}}
\caption{Left: KITTI frame. Centre: Generated fake CARLA frame. Right: Reconstructed KITTI frame.}
\label{fig:K2C}
\end{figure}

\begin{figure}
\centering
\fbox{\includegraphics[scale=0.27]{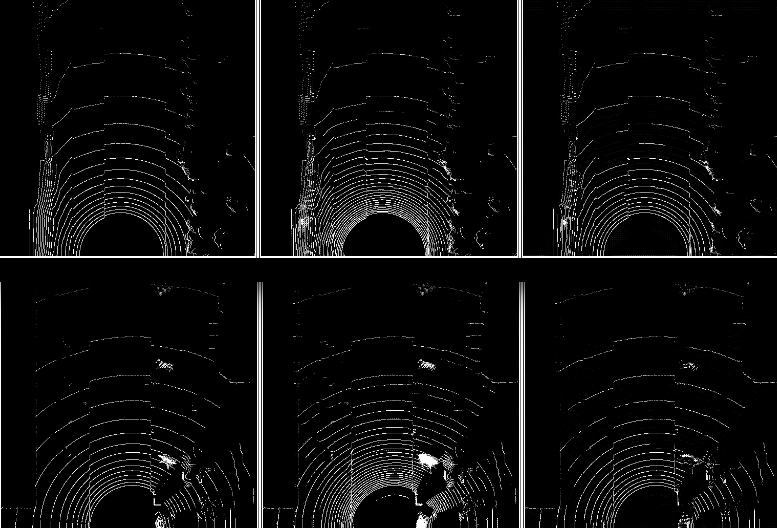}}
\caption{Left: CARLA frame with 32 channels. Middle: fake CARLA frame with 64 channels. Right: Reconstructed CARLA frame with 32 channels.}
\label{fig:32_64}
\end{figure}

\begin{figure}
\centering
\fbox{\includegraphics[scale=0.27]{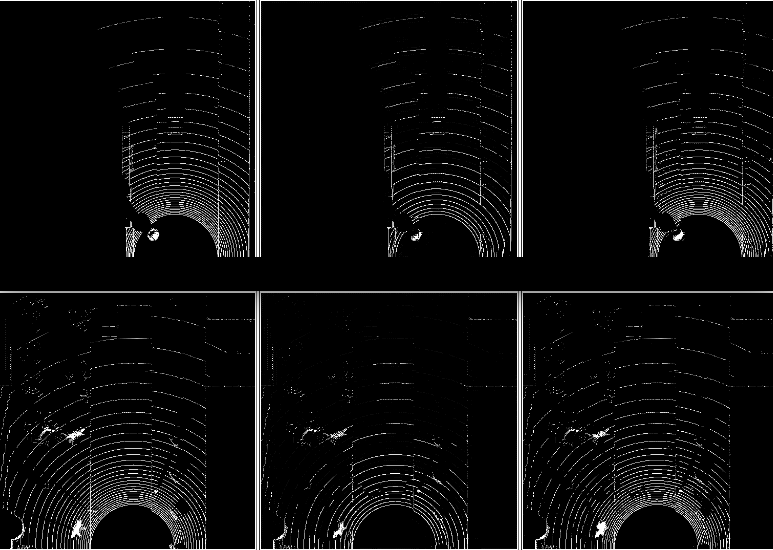}}
\caption{Left: CARLA frame with 64 channels. Middle: fake CARLA frame with 32 channels. Right: Reconstructed CARLA frame with 64 channels.}
\label{fig:64_32}
\end{figure}

\begin{figure}
\centering
\fbox{\includegraphics[scale=0.3]{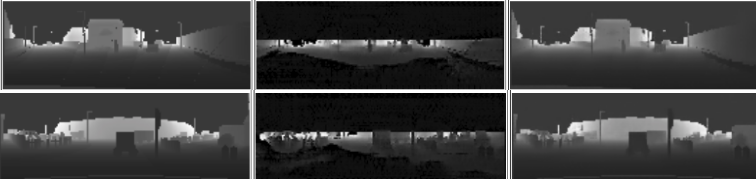}}
\caption{Left: Simulated CARLA PGM frame. Centre: Generated KITTI PGM frame. Right: Reconstructed CARLA PGM frame.}
\label{fig:PGM}
\end{figure}

\textbf{Sim2Real (BEV):}
The first experiment is to perform Image-to-Image LiDAR BEV translation from the CARLA simulator frames to the realistic KITTI frames and vice versa.
The number of training samples is 2000 frames in each domain. $\lambda$ of the Cycle consistency loss was set to 50.0 as it is noted that increasing this parameter helps in reconstructing images in both directions leading to generating better images.
The results of translation from simulated CARLA to realistic KITTI is shown in figure \ref{fig:C2K}. It is clear that the network is able to generate the corresponding KITTI fake BEV from the input CARLA frame while maintaining the general aspects of KITTI data and generating the objects found in the input image. As expected, the reconstruction at the left looks almost identical to the input. The opposite translation is shown in figure \ref{fig:K2C}, where the network is somewhat able to generate CARLA fake BEV. However, it did not maintain all the objects found in the input image. However, in this work, it is more important to translate from CARLA to KITTI.\\
\textbf{Real2Real (BEV):}
In another experiment, we test the translation between different sensors models configurations. In particular, we test the ability of CycleGAN to translate from a domain with few LiDAR channels to another domain with denser channels. The results of this translations are shown in figure \ref{fig:32_64}, \ref{fig:64_32}. It is shown that the network could translate the content from the 32 layers, but with the style of the dense 64 layers.\\
\textbf{Sim2Real (PGM):}
Finally, we test another projection method of LiDAR in the polar coordinate system, generating a Polar Grid Map (PGM). The way the LiDAR scans the environment is by sending multiple laser rays in the vertical direction, which define the number of LiDAR channels or layers. Each beam scans a plan, in the horizontal radial direction with a specific angular resolution. A LiDAR sensor configuration can be defined in terms of the number of channels, the angular resolution of the rays, in addition to its Field of View (FoV). PGM takes the $3$D LiDAR point cloud and fit it into a $2$D grid of values. The rows of the $2$D grid represent the number of channels of the sensor (For example, 64 or 32 in case of Velodyne). The columns of the $2$D grid represent the LiDAR ray step in the radial direction, where the number of steps equals the FoV divided by the angular resolution. The value of each cell is ranging from $0.0$ to $1.0$ describing the distance of the point from the sensor. PGM translations are shown in figure \ref{fig:PGM}. Such an encoding has many advantages. First, all the 3D LiDAR points are included in the map. Second, there is no memory overhead, like Voxelization for instance. Third, the resulting representation is dense, not sparse.


\begin{figure}
\centering
\fbox{\includegraphics[scale=0.41]{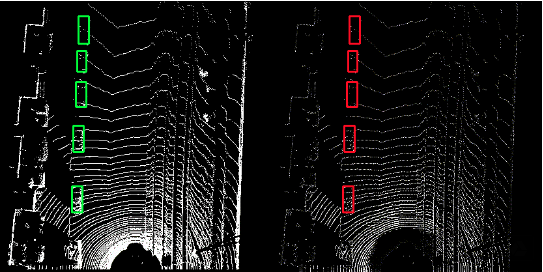}}
\caption{Annotation Transfer}
\label{fig:ann_transfer}
\end{figure}

\section{Discussion and future work}
\label{sec: Discussion}
\subsection{Evaluation of the generated LiDAR quality}
There are number of approaches for evaluating GANs. For example, Inception Score (IS)~\cite{salimans2016improved} provides a method to evaluate both the quality and diversity of the generated samples. IS is found to be well correlated with scores from human annotators. In Fr\'{e}chet Inception Distance (FID)~\cite{heusel2017gans}, the samples are embedded into a feature space and modeled as a continuous multivariate Gaussian distribution, where the Fr\'{e}chet distance is evaluated between the fake and the real data.
Thus far, the only way to evaluate the quality of the generated LiDAR is either subjective, through human judgment, or intrinsic, through the quality of the reconstruction in sim2real setup (CARLA2KITTI), or real2sim setup (KITTI2CARLA).
In this work, we present another method, which is "annotation transfer" in figure \ref{fig:ann_transfer}, by viewing the object bounding boxes from the annotations in one domain, and project it on the generated image in the new domain. Again, this method can serve as visual assessment, and also subject to human judgment.

\subsection{Extrinsic evaluation of the generated LiDAR quality}
\begin{figure}
\centering
\fbox{\includegraphics[scale = 0.31]{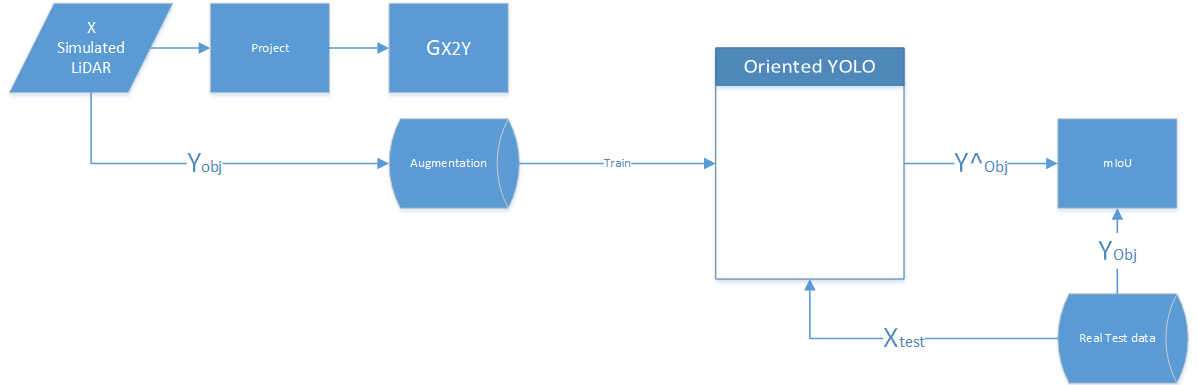}}
\caption{Extrinsic evaluation of the generated LiDAR PCL using Oriented YOLO}
\label{fig:gan_yolo}
\end{figure}
As a step towards a quantitative extrinsic assessment, one can think of using an external "judge," that evaluates the quality of the generated LiDAR in terms of the performance on a specific computer vision task, object detection for instance. The evaluation can be conducted using a standard benchmark data set $Y_{Obj}$ , which is manually annotated like KITTI. For that purpose, an object detection network from LiDAR is trained, For example \cite{ali2018yolo3d} and \cite{el2018yolo4d}. Starting from simulated data $X$, we can get the corresponding generated $Y$ data using $G$. We have the corresponding simulated data annotations $Y_{Obj}$. Object detection network can be used to obtain $\hat{Y}_{Obj}$, which can be compared in terms of mIoU to the exact $Y_{Obj}$. Moreover, Object detection network can be initially trained on the real benchmark data, KITTI, and then augment the data with the simulator generated data, and the mIoU can be monitored continuously to measure the improvement achieved by the data augmentation. The entire evaluation pipeline is shown in figure \ref{fig:gan_yolo}.

In essence, we want to ensure that, the essential scene objects are mapped correctly when transferring the style of the realistic LiDAR. This can be thought as similar to the content loss in the style transfer networks. To ensure this consistency, an extrinsic evaluation model is used. We call this model a reference model. The function of this model is to map the input LiDAR ($X$ or $Y$) to a space that reflects the objects in the scene, like semantic segmentation or object detection networks. For instance, this network could be YOLO network trained to detect objects from LiDAR views. More information is provided in the Appendix section \ref{Appendix}.

\subsection{Task specific loss}
In this part, we want to augment the cycle consistency losses $\mathcal{L}_{R_X}$ and $\mathcal{L}_{R_Y}$ with another loss that evaluates the object information existence in the mapped LiDAR from simulation to realistic.  

\section{Conclusions}
\label{sec: conclusion}
In this work, we tackled the problem of LiDAR sensor modeling in simulated environments, to augment real data with synthetic data. The problem is formulated as an image-to-image translation, where CycleGAN is used to map $sim2real$ and $real2real$. KITTI is used as the source of real LiDAR PCL, and CARLA as the simulation environment. The experimental results showed that the network is able to translate both BEV and PGM representations of LiDAR. Moreover, the intrinsic metrics, like reconstruction loss, and visual assessments show a high potential of the proposed solution. 
We discuss in the future works a method to evaluate the quality of the generated LiDAR, and also we presented an algorithm to extend the Vanilla CycleGAN with a task-specific loss, to ensure the correct mapping of the critical content while doing the translation. 
\bibliographystyle{icml2019}
\bibliography{nips_2018.bib}

\section{Appendix}
\label{Appendix}
\subsection{Extrinsic evaluation of the generated LiDAR PCL using Oriented YOLO}

We train two models:

1. A trained reference model for simulated LiDAR:
\begin{eqnarray}
\mathcal{L_{H_X}}(X, X^{GT}) \!\!\!\!\! &=& \!\!\!\!\! \mathbb{E}_{x \sim{~} p_{data}(x)}[T_X]\nonumber\\ 
 \!\!\!\!\! &= & \!\!\!\!\!\mathbb{E}_{x \sim{~} p_{data}(x)} \Big[|| H_X(x) - X^{GT} ||_2 \Big] \;, \nonumber \\ 
\end{eqnarray}

where $T_X = || H_X(x) - X^{GT} ||_2$
\begin{eqnarray}
H_X = \arg \min_H(\mathcal{L_{H_X}})
\end{eqnarray}

Where $X^{GT}$ is the ground truth data corresponding to the input $X$. 

For simulated data, this data is freely available by the simulator like CARLA.

2. A trained reference model for real LiDAR:
\begin{eqnarray}
\mathcal{L_{H_Y}}(Y, Y^{GT}) ) \!\!\!\!\! &=& \!\!\!\!\! \mathbb{E}_{y \sim{~} p_{data}(y)}\Big[|| H_Y(y) - Y^{GT} ||_2\Big] \nonumber\\ 
 \!\!\!\!\! &= & \!\!\!\!\! \mathbb{E}_{y \sim{~} p_{data}(y)}[T_Y]\;, \nonumber \\ 
\end{eqnarray}

where: $ T_Y = || H_Y(y) - Y^{GT} ||_2$

\begin{eqnarray}
H_Y = \arg \min_H(\mathcal{L_{H_Y}})
\end{eqnarray}

Where $Y^{GT}$ is the ground truth data corresponding to the input $Y$, real data requires human effort for manual annotation.

1. Evaluation of sim2real mapping using $H_Y$ model
\begin{eqnarray}
\hat{R_Y} = \Big[|| X^{GT} - H_Y(G(x))||_2\Big]
\end{eqnarray}

\begin{eqnarray}
\mathcal{L_{\hat{R_Y}}}(G, H_Y, X^{GT}, X) = \mathbb{E}_{x \sim{~} p_{data}(x)}\hat{R_Y}
\end{eqnarray}

Note that; here $X^{GT}$ is easy to obtain, but $H_Y$ is hard to train since it requires human labeling effort as described above.

2. Evaluation of real2sim mapping using $H_X$ model
\begin{eqnarray}
\hat{R_X} = \Big[|| Y^{GT} - H_X(F(y))||_2\Big]
\end{eqnarray}

\begin{eqnarray}
\mathcal{L_{\hat{R_X}}}(F, H_X, Y^{GT}, Y) = \mathbb{E}_{y \sim{~} p_{data}(y)}\hat{R_X}
\end{eqnarray}

Note that; $H_X$ is easy to train since it needs ground truth from simulator $X^{GT}$. However, $Y^{GT}$ is hard to obtain to evaluate the loss, since it is for real data, and hence requires human annotation.

\subsection{Supervised CycleGAN}

\begin{figure}
\centering
\fbox{\includegraphics[scale=0.16]{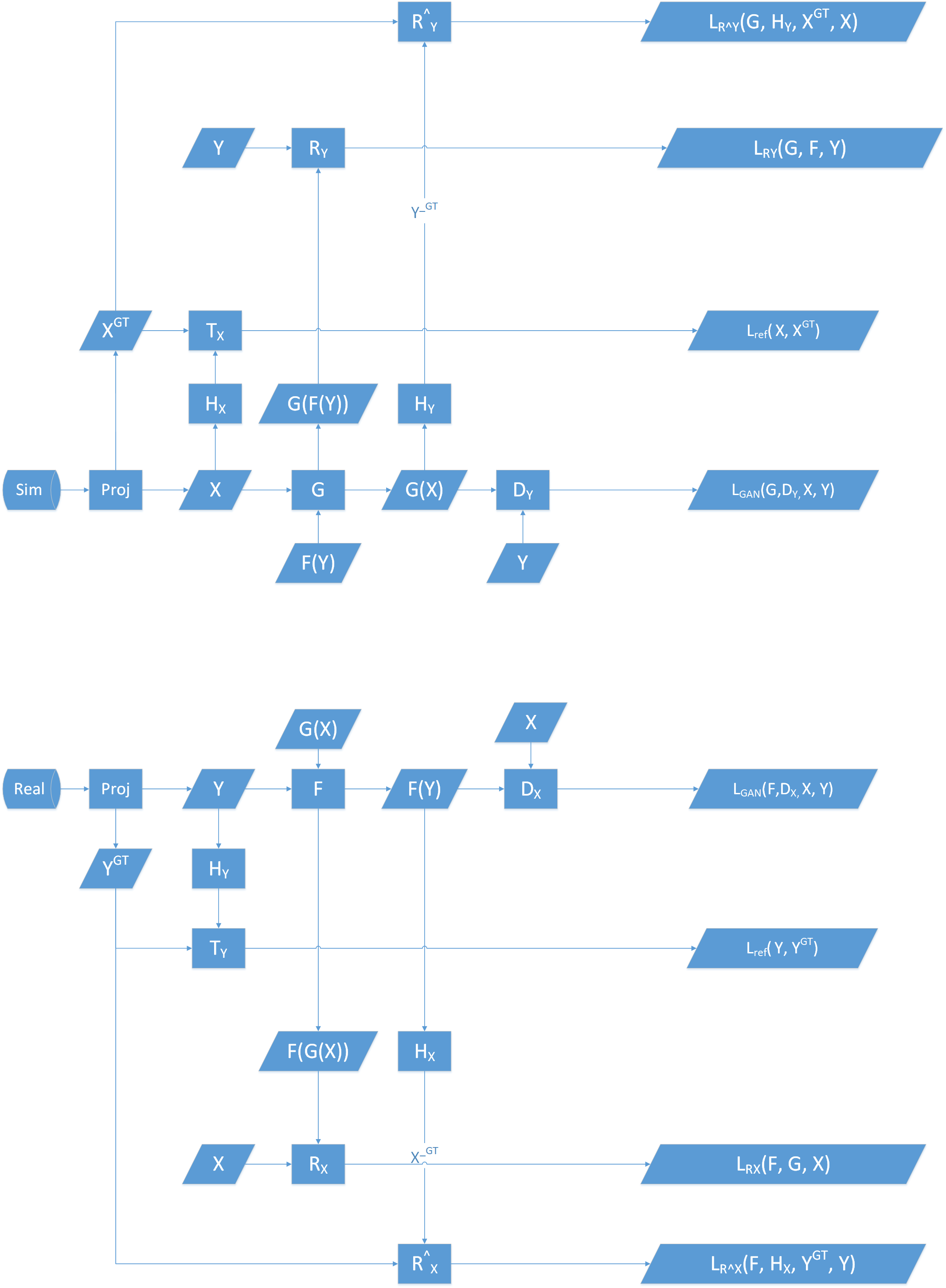}}
\caption{Adding object detection as task specific loss}
\label{fig:yolo_cyclegan}
\end{figure}

As you can see above, in both added loss terms, there's a difficulty related to getting real annotated data. Hence the proposed approach is limited in this regard. The limitation is relaxed if we already have a pre-trained network for the extrinsic task. 

The more importance is given to $H_Y$ network, and to the loss part $\mathcal{L_{\hat{R_Y}}}$, as it ensures the mapping $G$ preserves the object information in $X$. The part $\mathcal{L_{\hat{R_X}}}$ has less importance, as it relates to $F$, which we don't care much about.

\subsubsection{Advantage term}

The above two losses have a baseline performance that we cannot expect to exceed on average, which is the performance of $H_X$ and $H_Y$ on their training data, or $T_X$ and $T_Y$.

It is recommended to shift the losses by this advantage terms:
\begin{eqnarray}
\mathcal{L_{\hat{R_Y}}}(G, H_Y, X^{GT}, X) = \mathbb{E}_{x \sim{~} p_{data}(x)}[\hat{R_Y}] - T_Y
\end{eqnarray}

\begin{eqnarray}
\mathcal{L_{\hat{R_X}}}(F, H_X, Y^{GT}, Y) = \mathbb{E}_{y \sim{~} p_{data}(y)}[\hat{R_X}] - T_X
\end{eqnarray}

\subsubsection{Overall loss}

\begin{eqnarray}
\mathcal{L}(G,F,D_X,D_Y) \!\!\!\!\! &=& \!\!\!\!\! 
    \mathcal{L}_{GAN}(G, D_Y, X,Y) +\mathcal{L}_{GAN}(F, D_X, X,Y)\nonumber\\ 
 \!\!\!\!\! & & \!\!\!\!\! +\ \lambda_{cyc} \Big[\mathcal{L}_{R_Y}(G, F, Y)+\mathcal{L}_{R_X}(F, G, X)\Big]\nonumber\\ 
 \!\!\!\!\! & & \!\!\!\!\! +\ \lambda_{ref} \Big[\mathcal{L_{\hat{R_Y}}}(G, H_Y, X^{GT}, X)\nonumber\\ 
 \!\!\!\!\! & & \!\!\!\!\! +\ \mathcal{L_{\hat{R_X}}}(F, H_X, Y^{GT}, Y)\Big]\nonumber\\ 
 \!\!\!\!\! & & \!\!\!\!\! +\ \lambda_{ext} \Big[\mathcal{L_{H_X}}(X,X^GT)+\mathcal{L_{H_Y}}(Y,Y^GT)\Big]\;, \nonumber \\ 
\end{eqnarray}

\begin{eqnarray}
G^*, F^*, H^*_X, H^*_Y = \arg \min_{G, F, H_X, H_Y} \max_{D_X, D_Y} \mathcal{L}(G,F,D_X,D_Y)
\end{eqnarray}

\subsubsection{Loss decomposition}

Notice that the terms bounded by $\lambda_{ext}$ are the only function of $H_X$ and $H_Y$, so the optimization of $G$ and $F$ does not depend on those terms, and hence can be removed as follows:

\begin{eqnarray}
\mathcal{L}(G,F,D_X,D_Y) \!\!\!\!\! &=& \!\!\!\!\!
     \mathcal{L}_{GAN}(G, D_Y, X,Y) + \mathcal{L}_{GAN}(F, D_X, X,Y) \nonumber\\ 
     \!\!\!\!\! & & \!\!\!\!\! +\ \lambda_{cyc} \Big[\mathcal{L}_{R_Y}(G, F, Y) + \mathcal{L}_{R_X}(F, G, X)\Big] \nonumber\\ 
     \!\!\!\!\! & & \!\!\!\!\! +\ \lambda_{ref} \Big[\mathcal{L_{\hat{R_Y}}}(G, H_Y, X^{GT}, X)\nonumber\\ 
 \!\!\!\!\! & & \!\!\!\!\! +\ \mathcal{L_{\hat{R_X}}}(F, H_X, Y^{GT}, Y)\Big]\;, \nonumber \\ 
\end{eqnarray}

\begin{eqnarray}
 G^*, F^* = \arg \min_{G, F} \max_{D_X, D_Y} \mathcal{L}(G,F,D_X,D_Y)
\end{eqnarray}

On the other hand, $H_X$ and $H_Y$ only depend on the losses bounded by $\lambda_{ref}$ and $\lambda_{ext}$, hence we can update $H_X$ and $H_Y$ based on the losses above, only we rename $\lambda_{ref}$ (or add new scalar) into $\lambda_{aug}$:

\begin{eqnarray}
H_X  \!\!\!\!\! &=& \!\!\!\!\! \arg \min_H \Big(\lambda_{ext} \mathcal{L_{H_X}}(X,X^GT) \nonumber\\ 
 \!\!\!\!\! & & \!\!\!\!\! +\ \lambda_{aug} \mathcal{L_{\hat{R_Y}}}(G, H_Y, X^{GT}, X) \Big)\;, \nonumber \\ 
\end{eqnarray}

\begin{eqnarray}
H_Y \!\!\!\!\! &=& \!\!\!\!\! \arg \min_H \Big(\lambda_{ext} \mathcal{L_{H_Y}}(Y,Y^GT) \nonumber\\ 
 \!\!\!\!\! & & \!\!\!\!\! +\ \lambda_{aug} \mathcal{L_{\hat{R_X}}}(F, H_X, Y^{GT}, Y) \Big)\;, \nonumber \\ 
\end{eqnarray}

This is equivalent to a data augmentation scenario.

\textbf{Scenario (1)}: $\lambda_{ref} = \lambda_{aug} = \lambda_{ext} = 0$
This case is the Vanilla CycleGAN case

\textbf{Scenario (2)}: $\lambda_{ref} = 0, \lambda_{ext} = 0, \lambda_{aug} \neq 0$
This scenario is equivalent to using extrinsic evaluation $H_X$ and $H_Y$ for evaluation of the $G$ and $F$ mappings.

\textbf{Scenario (3)}: $\lambda_{ref} = 0, \lambda_{ext} \neq 0, \lambda_{aug} \neq 0$
This scenario is equivent to using extrinsic evaluation $H_X$ and $H_Y$ for evaluation of the $G$ and $F$ mappings, but augmenting all data (GT and Generated).

\textbf{Scenario (4)}: $\lambda_{ref} \neq 0,  \lambda_{aug} = 0, \lambda_{ext} = 0$
This case is equivalent to using frozen $H_X$ and $H_Y$, and in this case no data augmentation is happening.

\textbf{Scenario (5)}: $\lambda_{ref} \neq 0,  \lambda_{aug} \neq 0, \lambda_{ext} = 0$
This case is equivalent to updating $H_X$ and $H_Y$ with only generated data. This is like pre-training.

\textbf{Scenario (6)}: $\lambda_{ref} \neq 0,  \lambda_{aug} \neq 0, \lambda_{ext} \neq 0$
This is the general case.

\subsubsection{Algorithm based on Loss decomposition}

1. Train the reference networks: 
\begin{eqnarray}
H_X = \arg \min_H(\mathcal{L_{H_X}})
\end{eqnarray}

\begin{eqnarray}
H_Y = \arg \min_H(\mathcal{L_{H_Y}})
\end{eqnarray}

2. Freeze $H_X$ and $H_Y$:
\begin{eqnarray}
G^*, F^* = \arg \min_{G, F} \max_{D_X, D_Y} \mathcal{L}(G,F,D_X,D_Y)
\end{eqnarray}

3. Freeze $G$, $F$ and update $H_X$ and $H_Y$ based on the losses:
\begin{eqnarray}
H_X \!\!\!\!\! &=& \!\!\!\!\!\arg \min_H \Big(\lambda_{ext} \mathcal{L_{H_X}}(X,X^GT) \nonumber\\ 
 \!\!\!\!\! & & \!\!\!\!\! +\ \lambda_{aug} \mathcal{L_{\hat{R_Y}}}(G, H_Y, X^{GT}, X) \Big)\;, \nonumber \\ 
\end{eqnarray}

\begin{eqnarray}
H_Y \!\!\!\!\! &=& \!\!\!\!\! \arg \min_H \Big(\lambda_{ext} \mathcal{L_{H_Y}}(Y,Y^GT) \nonumber\\ 
 \!\!\!\!\! & & \!\!\!\!\! +\ \lambda_{aug} \mathcal{L_{\hat{R_X}}}(F, H_X, Y^{GT}, Y) \Big)\;, \nonumber \\ 
\end{eqnarray}
Here we can have settings from Scenario 3 (pre-training) or 4 (augmentation).

\end{document}